\documentclass{article}
\PassOptionsToPackage{numbers, compress}{natbib}

\usepackage[preprint]{neurips_2023}

\usepackage[utf8]{inputenc} 
\usepackage[T1]{fontenc}    
\usepackage{xcolor}         
\usepackage[colorlinks]{hyperref}      
\usepackage{url}            
\usepackage{booktabs}       
\usepackage{amsfonts}       
\usepackage{amsmath}
\usepackage{mathtools}      %
\usepackage{nicefrac}       
\usepackage{microtype}      
\usepackage[pdftex]{graphicx}
\usepackage{caption}
\usepackage{subcaption}
\usepackage{wrapfig}
\usepackage{multirow}
\usepackage{verbatim}
\usepackage{algorithm}
\usepackage{enumitem}
\usepackage{comment}

\usepackage{xcolor,colortbl}
\usepackage{tcolorbox}

\usepackage{algorithmicx}
\usepackage{algorithm}
\usepackage[noend]{algpseudocode}
\algnewcommand{\LineComment}[1]{\State \(\triangleright\) #1} 
\setlist{noitemsep} 
\setlist[itemize]{leftmargin=*}
\usepackage[export]{adjustbox}
\makeatletter
\g@addto@macro{\UrlBreaks}{\UrlOrds}
\makeatother

\definecolor{light-gray}{gray}{1.0}
\newcommand{\code}[1]{{\texttt{#1}}}
\newcommand{\blockcomment}[1]{}
\makeatletter
\newcommand\footnoteref[1]{\protected@xdef\@thefnmark{\ref{#1}}\@footnotemark}
\newcommand*{\rom}[1]{\expandafter\@slowromancap\romannumeral #1@}
\makeatother
\DeclareMathAlphabet\mathbfcal{OMS}{cmsy}{b}{n}


\title{RepoFusion: Training Code Models to Understand Your Repository}

\author{%
  Disha Shrivastava\thanks{Correspondence to: <dishu.905@gmail.com>}\\
  Mila, Universit\'e de Montr\'eal
  \And
  Denis Kocetkov\\
  ServiceNow Research
  \And 
  Harm de Vries\\
  ServiceNow Research
\And
  Dzmitry Bahdanau\\
  ServiceNow Research\\
  Mila, McGill University\\
  Canada CIFAR AI Chair
\And
  Torsten Scholak\\
  ServiceNow Research
}

\begin{document}

\maketitle

\begin{abstract}
Despite the huge success of Large Language Models (LLMs) in coding assistants like GitHub Copilot~\footnote{\label{copliot}https://github.com/features/copilot/}, these models struggle to understand the context present in the repository (e.g., imports, parent classes, files with similar names, etc.), thereby producing inaccurate code completions. This effect is more pronounced when using these assistants for repositories that the model has not seen during training, such as proprietary software or work-in-progress code projects. Recent work~\cite{shrivastava2022repository, repocoder} has shown the promise of using context from the repository during inference. In this work, we extend this idea and propose \emph{RepoFusion}, a framework to train models to incorporate relevant repository context. Experiments on single-line code completion show that our models trained with repository context significantly outperform much larger code models as CodeGen-16B-multi ($\sim73\times$ larger) and closely match the performance of the $\sim 70\times$ larger StarCoderBase model that was trained with the Fill-in-the-Middle objective. We find these results to be a novel and compelling demonstration of the gains that training with repository context can bring. We carry out extensive ablation studies to investigate the impact of design choices such as context type, number of contexts, context length, and initialization within our framework. Lastly, we release \emph{Stack-Repo}, a dataset of 200 Java repositories with permissive licenses and near-deduplicated files that are augmented with three types of repository contexts. Additionally, we are making available the code and trained checkpoints for our work. Our released resources can be found at: \url{https://huggingface.co/RepoFusion}.
\end{abstract}

\section{Introduction}\label{intro}

Large Language Models (LLMs) of code~\cite{svyatkovskiy2020intellicode, codex, incoder, code-t5, codegen, alphacode, santacoder} have gained significant popularity. The demand for these models has further increased with their integration into code assistants like GitHub Copilot~\footnoteref{copliot} and TabNine~\footnote{https://www.tabnine.com/}, and their popularity is anticipated to grow further as more developer-assistance products are developed around them.

Despite their remarkable capabilities, LLMs of code often struggle to generalize effectively in unforeseen or unpredictable situations, resulting in undesirable predictions. Instances of such scenarios include code that uses private APIs or proprietary software, work-in-progress code, and any other context that the model has not seen while training. To address these limitations, one possible approach is to enhance the predictions of these models by incorporating the wider context available in the repository. Leveraging the structure and context of the repository can take into consideration dependencies between files, such as imports and parent classes, and provide valuable insights into coding patterns that may be specific to the organization or user. Recent works~\cite{shrivastava2022repository, repocoder, cocomic} have shown promising results in utilizing repository-level context in conjugation with LLMs of code. It was also shown in ~\citet{shrivastava2022repository} that without specialized training, it is challenging to integrate multiple relevant contexts from the repository. Building upon these findings we propose RepoFusion, a training framework for learning to combine multiple relevant contexts from the repository in order to generate more accurate and context-aware code completions. 

\begin{figure}[t]
  \centering
  \includegraphics[width=1.0\textwidth]{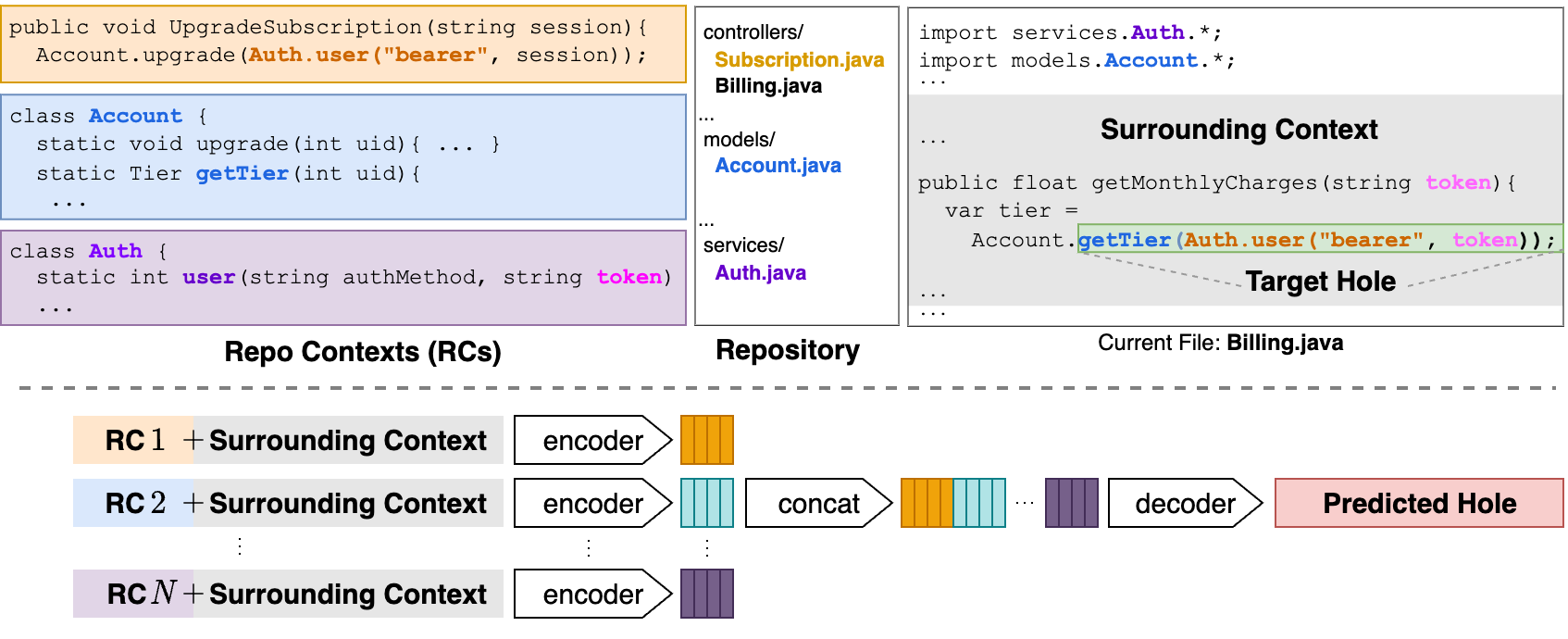}
  \caption{\textbf{Figure explaining the idea of RepoFusion}. Given multiple relevant contexts from the repository (Repo Contexts), RepoFusion appends the Surrounding Context (highlighted in gray) to each repo context, encodes them separately, and combines them to produce a prediction of the target hole (highlighted in red).}
  \label{fig:block_diagram}
\end{figure}

In this work, we focus on the task of single-line code completion~\cite{n-gram, shrivastava2020onthefly} which simulates real-world scenarios where users are editing existing files in an IDE. With reference to Figure~\ref{fig:block_diagram}, this means that we have to predict the missing section, referred to as the \emph{target hole} (highlighted in green), starting from the cursor's position until the end of the line. We see that the completion of this target hole will benefit from context not just in the current file (the variable name \code{token}), but also from other files in the repository. Specifically, the context from the imported file Account.java provides insight into the usage of the \code{getTier} method, while the sibling file Subscription.java offers guidance on the usage of \code{Auth.user("bearer"}, with the definition of \code{Auth} found in the imported file Auth.java. Given these relevant code snippets from across the repository which we call \emph{Repo Contexts (RCs)}, RepoFusion uses the Fusion-in-Decoder~\cite{fid} architecture to combine these. Specifically, each repo context is appended with the \emph{surrounding context} i.e., a window around the target hole excluding the target hole (highlighted in gray) and encoded separately. A decoder jointly attends to the concatenated encoded representations to produce a prediction for the target hole (highlighted in red). The key contributions of our paper can be listed as follows:
\begin{itemize}
    \item We propose RepoFusion, a framework that helps code models to make better predictions by learning to combine relevant contextual cues from the repository.
    \item Through extensive experiments we establish that RepoFusion, a 220M parameter model, significantly outperforms several larger models trained on the next-token prediction objective such as CodeGen-16B~\cite{codegen}. Furthermore, despite being approximately 70 times smaller in size, our model closely matches the performance of StarCoderBase~\cite{starcoder}, a 15.5B parameter LLM trained with the Fill-in-the-Middle~\cite{bavarian2022efficient} objective.
    \item We conduct thorough ablation studies to gain insights into the key factors influencing RepoFusion, such as the nature of repository contexts, their lengths, the number of repository contexts, and other training configurations. One of the crucial findings is that leveraging information from diverse sources within a repository is the key to RepoFusion's effectiveness.
    \item We create and release \href{https://huggingface.co/datasets/RepoFusion/Stack-Repo}{\emph{Stack-Repo}}, a dataset of 200 Java repositories with permissive licenses and near-deduplicated files that are augmented with three types of repository contexts. Our released resources can be found at: \url{https://huggingface.co/RepoFusion}.
\end{itemize}

\section{Training with Repository Context}
In this section, we briefly describe Fusion-in-Decoder~\cite{fid}, the repository contexts we used, and the details of our RepoFusion framework.

\subsection{Fusion-in Decoder} \label{fid}
Fusion-in-Decoder~\cite{fid}~(FiD) is a method to train a language model to combine information coming from multiple sources. In the original work, FiD was used for open-domain question answering. In the FiD approach to question answering, a sequence-to-sequence model takes support passages concatenated with the question as inputs and produces an answer as the output. Each support passage is appended to the question and encoded independently by the encoder. The encoded representations are then concatenated and fed to the decoder which jointly attends to them to produce the answer. In this work, we adapt FiD for the setting of code completion.

\subsection{Repository Contexts}\label{repo_context}
Motivated by the syntax and semantics of programming languages as well as the common coding patterns, \citet{shrivastava2022repository} proposed a set of repo-level prompt proposals that leverage the structure and the relevant context in files across the repository. 
A prompt proposal (PP) is a function that takes in the target hole's location and the associated repository as input and returns a string called \emph{Prompt Proposal Context (PPC)} as output. The prompt proposal context is created by extracting a particular type of context (prompt context type) from a particular category of related source files (prompt source). Examples of prompt sources are the current file, files that are imported into the current file, files that have a similar name as the current file, etc. Examples of prompt context types are lines following the target hole, method names and bodies, identifiers, string literals, etc. Combining these prompt sources and prompt context types gives us a total of 63 prompt proposals (see Appendix B.4 of \citet{shrivastava2022repository} for details). It should be noted that the context from the beginning of the current file up to the position of the hole, as well as the context following the target hole within the current file are also types of prompt proposal contexts. We will refer to these as the prior PPC (or just \emph{prior}) and the post PPC (or just \emph{post}), respectively in the remainder of the paper. Note that depending on the target hole, some prompt proposal contexts might be empty (e.g. if the target hole is towards the very beginning of the file, there might not be any import statements from the current file to get context from). 

Repo-level prompt proposals can be thought of as a deterministic retrieval mechanism that returns the relevant code snippets from the repository. To understand the role of the retrieved repo contexts, apart from prompt proposals, we also consider two other mechanisms for retrieving repository-level context (see Appendix~\ref{app:implementation_retrieval_mechanisms} for implementation details): (a) BM25: The context from each file in the repository is scored using BM25-based~\cite{bm25} similarity with the surrounding context, and (b) RandomNN (also used in \citet{shrivastava2022repository}): From a list of randomly selected chunks from the repository, we select the top-k based on the similarity of the embedded chunks with the embedded surrounding context in the representation space. 

\subsection{RepoFusion}\label{repofusion}
\begin{figure}[t]
  \centering
  \includegraphics[width=1.0\textwidth]{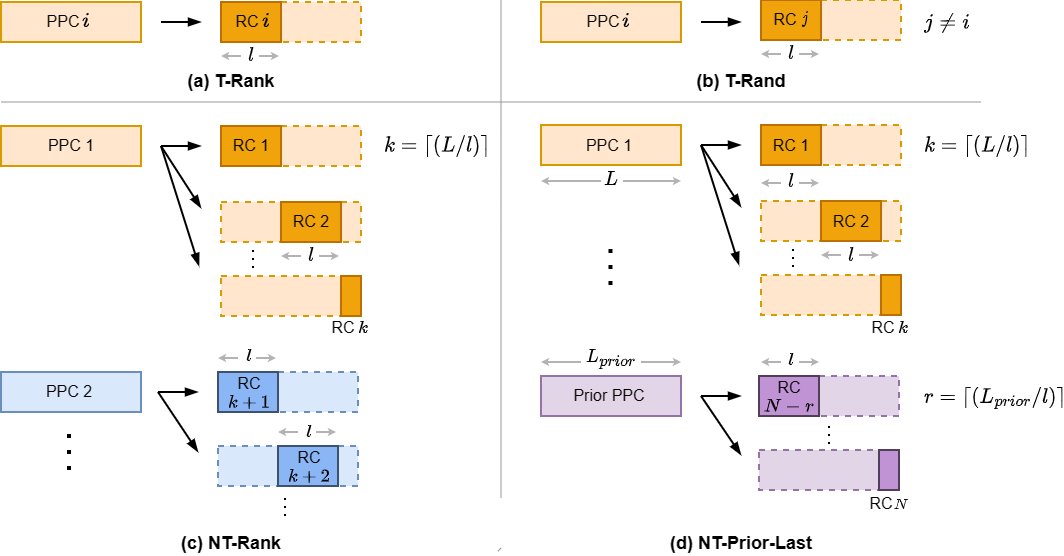}
  \caption{\textbf{Different strategies employed for producing repo contexts (RCs) from prompt proposal contexts (PPCs)}: 
  \textbf{(a) T-rank:} we truncate the $i$-th ranked PPC to yield the $i$-th RC. \textbf{(b) T-rand:} we position the truncated $i$-th PPC at a random position $j$ in RepoFusion's sequences of RCs. \textbf{(c) NT-Rank:} each PPC yields as many RCs as necessary to exhaust all of its tokens without truncation. \textbf{(d) NT-Prior-Last:} we reserve the last $r$ RCs for the Prior PPC and fill the rest RCs as in NT-Rank.}
  \label{fig:repo_context_strategies}
\end{figure}

The core idea of RepoFusion is to train a code model to be aware of the context in the repository such that it helps in generating an accurate prediction of the target hole. Given a set of retrieved repo contexts, RepoFusion learns to combine the relevant parts of these contexts using the  FiD approach as described in Section~\ref{fid}. The surrounding context is concatenated with each repo context and then encoded independently (see Figure \ref{fig:block_diagram}, bottom). Note that for our purpose, since we want the code model to complete the target hole, we append the surrounding context toward the end of the repo context. This is different from the original work~\cite{fid}, where the question (analogous to the surrounding context in our setting) is appended at the beginning of each passage (analogous to the repo context in our setting). RepoFusion uses $N$ repo contexts of length $l$ tokens each. We experimented with the following four strategies for producing and ordering the repo contexts based on the prompt proposal contexts (see Figure~\ref{fig:repo_context_strategies}).
\begin{enumerate}
    \item \textbf{Truncated-Ranked (T-Rank):} In this setting, one prompt proposal context yields one repo context. We truncate each prompt proposal context (i.e., take only the first $l$ tokens) to form the respective repo context and discard the rest. The repo contexts are ordered based on the ranking of the prompt proposals~\footnote{https://github.com/shrivastavadisha/repo\_level\_prompt\_generation/blob/main/get\_info\_from\_hole\_predictions.py} on the validation split of the Google Code archives dataset of \cite{shrivastava2022repository}. Given that our work and ~\citet{shrivastava2022repository} both target Java, it seemed reasonable to us to directly use this ordering.

    \item \textbf{Truncated-Random (T-Rand):} Same as T-rank except that the repo contexts are ordered randomly. This helps us understand the role of the specific ranking of PPs from \cite{shrivastava2022repository}.
    
    \item \textbf{Not Truncated-Ranked (NT-Rank):}     The prompt proposals are ranked based on the same order as in T-Rank. Unlike T-rank, here we avoid the truncation of prompt proposal contexts. Instead, we construct as many repo contexts from each prompt proposal context as necessary, namely a PPC of length $L$ will contribute $k = \lceil(L/l)\rceil$ RCs. We then proceed to the next in order prompt proposal and continue so until we have selected $N$ repo contexts. Unlike T-rank, this setting allows RepoFusion to see the entirety of top-ranked prompt proposals at the cost of potentially ignoring the lower-ranked ones.

    \item \textbf{Not Truncated-Prior-Last (NT-Prior-Last):} Same as NT-Rank except that the prior PPC is always ordered at the end. Since the decoder attends to the concatenated encoded representations of the repo contexts in the same order as it is presented as inputs, this strategy helps us understand the role of continuing code generation from the encoded representation of the prior PPC as the most recently attended representation in the decoder. Note that depending on the value of $N$, it may be necessary to remove certain chunks of top-ranked PPCs in order to accommodate the prior PPC at the end. 
\end{enumerate}

Similar to~\citet{fid}, we format each repo context with special tokens to mark the beginning of the surrounding context and the repo context, as well as for the name of the repo context (which is the same as the name of the prompt proposal). Please see the Appendix~\ref{app:implementation} for details on these tokens and other architectural details of RepoFusion. 

\section{Experiments and Results}\label{experiments}
In this section, we describe the process of creation of our dataset Stack-Repo and the details of experiments. We then present the results of evaluating RepoFusion and other models on the test set. This is followed by presenting the findings of extensive ablation studies carried out on the validation set to gain deeper insights into the individual contributions of each component in our framework.

\subsection{Dataset Creation}\label{dataset_creation} 
In this work, we build upon a modified version of The Stack V1.1~\cite{kocetkov2022stack}. The modified version~\footnote{https://huggingface.co/datasets/bigcode/the-stack-dedup} consists of near-deduplicated code repositories with permissive licenses from GitHub. For our experiments, we take only the Java subset (files with .java extension) of this dataset.

\textbf{Creation of Target Holes:} For creating target holes needed for training and evaluating RepoFusion, we choose a set of repositories randomly from the Java subset of the Stack and divide them into training, validation, and test splits in the ratios 2:1:1 respectively. We only consider repositories that contain at least 20 near-deduplicated files. For each repository, we choose target holes from every code line (excluding comments in natural language and blanks) in all the files. In order to tokenize a code line, we used common Java code delimiter tokens~\footnote{$\left[\code{.}, \code{(}, \code{)}, \code{[}, \code{]}, \code{ }, \code{\{}, \code{\}}, \code{,}, \code{:}, \code{"}, \code{;}\right]$}. 
We chose a random token within the line and the rest of the line starting from that position till the end constitutes the target hole. By not selecting target holes based on the tokenizer of a specific code model, we can ensure that the tokenizer remains unbiased and does not implicitly favor any particular code model in our experiments. To avoid bias from large repositories while training, we cap the maximum contribution of target holes from a repository to 10000, i.e.\ if the total number of holes in the repository exceeds 10000, we select 10000 holes randomly from the total holes. Please see Table~\ref{tab:dataset_stats} for the statistics of Stack-Repo. 

\begin{table}[H]
\centering
\caption{Statistics of Stack-Repo}
\scalebox{0.9}{
\begin{tabular}{lccc}
\toprule
\textbf{Feature} & \textbf{Train} & \textbf{Val} & \textbf{Test}\\
\midrule
\textbf{\# Repositories} & 100 & 50 & 50\\   
\textbf{\# Files} & 20310  & 11172 & 13202 \\
\textbf{\# Holes} & 435890 & 220615 & 159822\\
\bottomrule
\end{tabular}
\label{tab:dataset_stats}}
\end{table}

\textbf{Creation of Repo Contexts:} For each target hole, we use the implementation~\footnote{https://github.com/shrivastavadisha/repo\_level\_prompt\_generation} from \cite{shrivastava2022repository} to extract prompt proposal contexts. We take two lines above and two lines below the target hole excluding the target hole as the surrounding context. For obtaining the embeddings for RandomNN repo contexts, we use pre-trained CodeBERT~\cite{feng2020codebert}. For constructing the BM25 repo contexts, we use the implementation from the Rank-BM25 package~\footnote{https://pypi.org/project/rank-bm25/}. To improve efficiency, we store the repo contexts for each target hole in advance. Note that even though our target hole and repo context creation strategies have been inspired from \citet{shrivastava2022repository}, our dataset, Stack-Repo is significantly bigger in size. 
Apart from code completion, Stack-Repo can serve as a benchmark for various other code-related tasks involving repository context, such as bug repair and pull request resolution. We plan to release it under the same license as The Stack~\cite{kocetkov2022stack} to support future research in these areas. 

\subsection{Experimental Details}

\textbf{Training of RepoFusion:} We use the 220M parameter CodeT5-base~\cite{code-t5} encoder-decoder model as our base code model for RepoFusion. We found that the pre-trained CodeT5 model was not good at completing Java code (see Appendix~\ref{app:pretrained_codet5} for initial results). 
Therefore, to obtain a base model for RepoFusion training we finetuned CodeT5-base with an input context length of 512 using the next-token prediction objective on Java repositories from the dataset described in Section~\ref{dataset_creation}. Specifically, we used the repositories that were not included in Stack-Repo. For each file, we randomly sample ten pivot points with the code context prior to the pivot location in the file serving as the input to the encoder of CodeT5. 
The finetuned CodeT5-base model was then used to initialize the training of RepoFusion. 
Based on the validation set performance, we found that for RepoFusion, NT-Prior-Last with $N=32$ and $l=768$ works the best. We provide complete details of training RepoFusion and finetuning CodeT5 in Appendix~\ref{app:implementation}.

\textbf{Baselines:} \label{baselines}To benchmark the performance of RepoFusion, we conducted experiments with several methods, with each model utilizing the recommended tokenizers specific to the method and employing a maximum token generation limit of 128 per completion. To ensure a thorough analysis, we have incorporated encoder-decoder models as well as decoder-only models of different sizes, with varying context lengths and two different input context types. We present the details of the methods below:

\begin{enumerate}
    \item \textbf{CodeT5 (FT)}: In addition to the previously described fine-tuned (FT) version of \textbf{CodeT5-base}, we also finetuned \textbf{CodeT5-large} (770M) with a context length of 512. Next, we assessed the performance of these models using input context lengths of 2048 and 4096. The input context was constructed by either considering the prior PPC (\emph{prior}) alone or by concatenating equal lengths of the post PPC (\emph{post}) and prior. 
    
    \item \textbf{BigCode models:} We experimented with two models released by BigCode~\footnote{https://www.bigcode-project.org/}. The first model is \textbf{SantaCoder}~\cite{santacoder}, which is a 1.1B parameter model which supports a maximum context length of 2048 tokens and the second is the recently released \textbf{StarCoderBase}~\cite{starcoder} model which is a 15.5B parameter model that can support up to 8192 tokens. Both of these models are trained with Fill-in-the-Middle~\cite{bavarian2022efficient} (FiM) objective on versions 1.1 and 1.2 of The Stack~\cite{kocetkov2022stack}, respectively. These models were evaluated with both the prior and post+prior contexts as inputs. For experiments with post+prior, we used the FiM special tokens that were used while training these models. Since these models have been trained specifically to see the post PPC as suffix, they help us understand the role of training with multiple repo contexts in the way proposed by RepoFusion.
    
    \item \textbf{CodeGen~\cite{codegen}:} CodeGen is a decoder-only transformer-based autoregressive model trained with the next-token prediction objective. It supports a maximum context length of 2048 tokens. We experimented with three pre-trained variants of CodeGen, namely \textbf{CodeGen-2B-multi}, \textbf{CodeGen-6B-multi}, and \textbf{CodeGen-16B-multi}. As before, we tried the scenarios where the input context consists of the post + prior as well as when the input context consists of just the prior. These models help us understand the performance of large pre-trained models that are not trained with repo context. 
\end{enumerate}
It is important to note that when compared to our RepoFusion model, with the exception of CodeT5-base (FT), all other models are many times larger in size and have been trained on a significantly larger number of tokens. The rationale behind selecting these baselines is to compare the performance of training smaller models with additional repository context against training much larger models without incorporating repository context.

\textbf{Evaluation Metric:} We conduct an exact string match between the predicted hole and the target hole, where the predicted hole is the string up to the occurrence of the first newline in the completion. If an exact match is found, it is considered a success; otherwise, it is deemed a failure. We measure the fraction of exact matches over the dataset and call it \emph{Success Rate}.

\subsection{Results}
Table~\ref{tab:test_performance} presents the hole completion success rate (along with standard error) in percentage for different methods on our test set, where the standard error is an estimate of the variability in the sample mean of the distribution of exact match. The top two sections of the table display the evaluation results of the finetuned encoder-decoder \{CodeT5-base(FT), CodeT5-large(FT)\} models and decoder-only \{SantaCoder, CodeGen-2B, CodeGen-6B, CodeGen-16B\} models, respectively when provided with prior context as input. The table's next two sections present the results of evaluating these models when given with post+prior context as input.

In the final section of the table, we showcase the evaluation results of RepoFusion using different effective input context lengths obtained by varying the values of $N$ and $l$.

\begin{table}[t]
\centering
\caption{Completion success rate on the test set for different methods.}
\scalebox{1.0}{
\begin{tabular}{@{}cccccc@{}}
\toprule
\textbf{\begin{tabular}[c]{@{}c@{}}Model\end{tabular}} & 
\textbf{\begin{tabular}[c]{@{}c@{}} Size \\(\#params)\end{tabular}} &
\textbf{\begin{tabular}[c]{@{}c@{}} Effective \\ context length \end{tabular}} &
\textbf{\begin{tabular}[c]{@{}c@{}} Context \\ type \end{tabular}} &
\textbf{\begin{tabular}[c]{@{}c@{}} Success Rate \\(\%)\end{tabular}} \\
\midrule 
CodeT5-base (FT) & 0.22B & 2048 & prior & 41.82 $\pm$ 0.12\\ 
CodeT5-base (FT) & 0.22B & 4096 & prior & 46.45 $\pm$ 0.12\\
CodeT5-large (FT) & 0.77B & 2048 & prior & 44.73 $\pm$ 0.12\\
CodeT5-large (FT) & 0.77B & 4096 & prior & 48.92 $\pm$ 0.12\\
\midrule
SantaCoder & 1.1B & 2048 & prior &  39.51 $\pm$ 0.12\\
CodeGen & 2B & 2048 & prior &  49.45 $\pm$ 0.12\\ 
CodeGen & 6B & 2048 & prior &  49.19 $\pm$ 0.12\\ 
CodeGen & 16B & 2048 & prior & 50.20 $\pm$ 0.12\\
\midrule
CodeT5-base (FT) & 0.22B & 2048 & post+prior & 48.89 $\pm$ 0.12\\ 
CodeT5-base (FT) & 0.22B & 4096 & post+prior & 49.97 $\pm$ 0.12\\ 
CodeT5-large (FT) & 0.77B & 2048 & post+prior & 51.72 $\pm$ 0.12\\ 
CodeT5-large (FT) & 0.77B & 4096 & post+prior & 52.43 $\pm$ 0.12\\ 
\midrule
SantaCoder & 1.1B & 2048 & post+prior &  56.78 $\pm$ 0.12\\
CodeGen & 2B & 2048 & post+prior & 53.18 $\pm$ 0.12\\
CodeGen & 6B & 2048 & post+prior &  54.03 $\pm$ 0.12\\
CodeGen & 16B & 2048 & post+prior &  54.09 $\pm$ 0.12\\
\midrule
RepoFusion ($N=4$, $l=512$) & 0.22B & 2048 & NT-Prior-Last &  65.96 $\pm$ 0.12\\
RepoFusion ($N=8$, $l=512$) & 0.22B & 4096 & NT-Prior-Last &  70.38 $\pm$ 0.11\\
RepoFusion ($N=32$, $l=768$) & 0.22B & 24576 & NT-Prior-Last &  77.32 $\pm$ 0.10\\
\bottomrule
\end{tabular}}
\label{tab:test_performance}
\end{table}
\begin{table}[t]
\centering
\caption{Comparison with StarCoderBase on a test set subset.}
\scalebox{1.0}{
\begin{tabular}{@{}ccccc@{}}
\toprule
\textbf{\begin{tabular}[c]{@{}c@{}}Model\end{tabular}} & 
\textbf{\begin{tabular}[c]{@{}c@{}} Size \\(\#params)\end{tabular}} &
\textbf{\begin{tabular}[c]{@{}c@{}} Effective\\ context length \end{tabular}} &
\textbf{\begin{tabular}[c]{@{}c@{}} Context \\ type \end{tabular}} &
\textbf{\begin{tabular}[c]{@{}c@{}} Success Rate \\(\%)\end{tabular}}\\ 
\midrule
StarCoderBase & 15.5B & 8192 & prior & 52.97 $\pm$ 0.45\\ 
\midrule
StarCoderBase & 15.5B & 8192 & post+prior & 79.79 $\pm$ 0.36\\ 
\midrule
RepoFusion ($N=16$, $l=512$) & 0.22B & 8192 & NT-Prior-Last & 73.67 $\pm$ 0.43\\
RepoFusion ($N=32$, $l=2500$) & 0.22B & 80000 & NT-Prior-Last & 78.33 $\pm$ 0.37\\
\bottomrule
\end{tabular}
\label{fig:large_test_perf}}
\end{table}

\textbf{Baseline Performance Improves with Model Size and the Addition of Context:}
The performance of CodeT5 (FT) models improves as the model becomes bigger (CodeT5-large vs CodeT5-base) and as the input context length increases (2048 vs 4096). We observe a comparable pattern with decoder-only models, where there is a general enhancement in performance as the models grow larger (with a slight indication of saturation) while maintaining a fixed context length. Additionally, we note a substantial improvement in both categories of models when provided with post + prior context as input, compared to their respective performances with only the prior context. The SantaCoder model, specifically trained for the FiM task, exhibits the most significant improvement. 

\textbf{RepoFusion is Effective:}
RepoFusion not only exhibits a substantial improvement over its base model (CodeT5-base (FT)) but also surpasses other bigger models, even when utilizing the same effective context length. Furthermore, RepoFusion achieves superior performance compared to the significantly larger CodeGen-16B model, even when constrained to utilize fewer repository contexts to match the effective context length of CodeGen-16B. Furthermore, when provided with additional repo contexts, RepoFusion demonstrates further enhancements in performance. 

We also compare RepoFusion with the recently released StarCoderBase~\cite{starcoder} model. StarCoderBase is a 15.5B parameter model which is trained with about one trillion tokens using a FiM objective employing a large input context length of 8192 tokens. The results of this comparison using a random subset of 12500 holes from our test set are depicted in Table~\ref{fig:large_test_perf}. Learning to read additional repository context allows RepoFusion to achieve success rate just 1.3\% below the performance of the 70 times bigger state-of-the-art StarCoderBase model.

\begin{figure}[H]
\begin{subfigure}{.49\textwidth}
\includegraphics[width=1.0\linewidth]{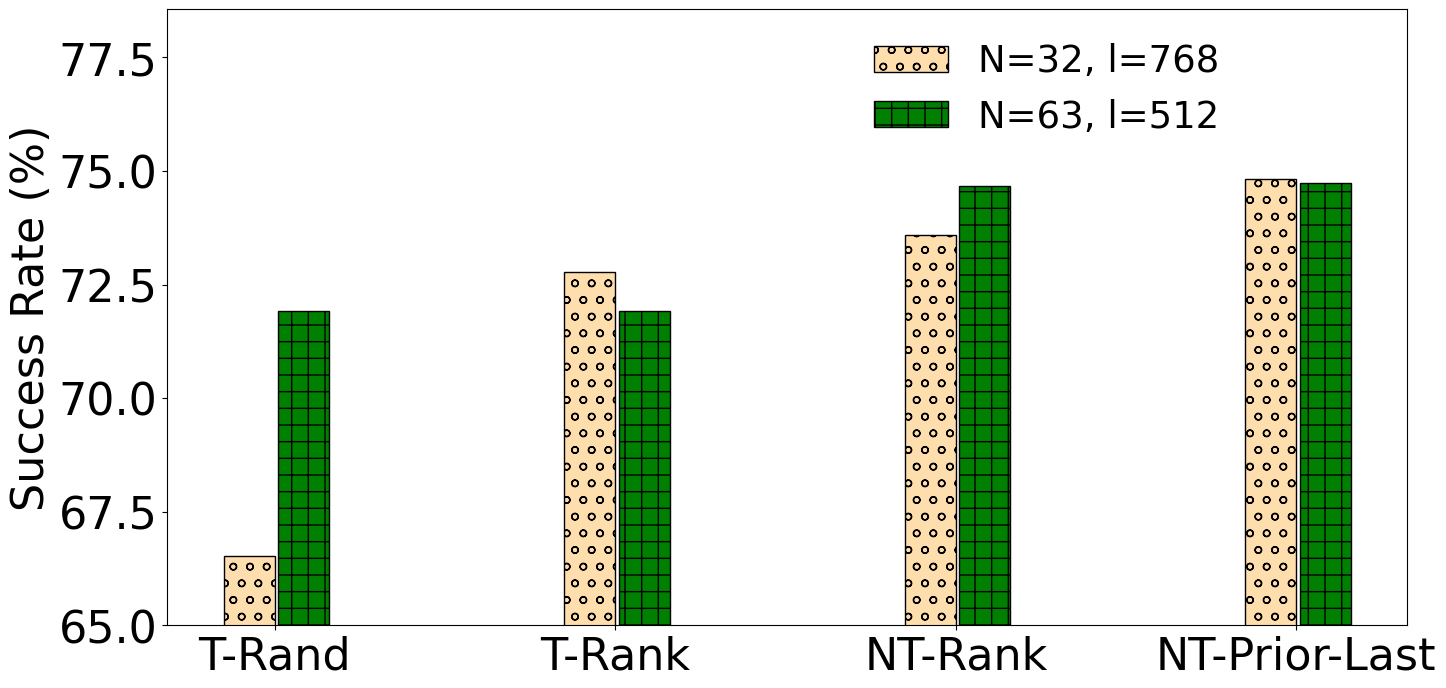}
\end{subfigure}
\hfill
\begin{subfigure}{.49\textwidth}
\includegraphics[width=1.0\linewidth]{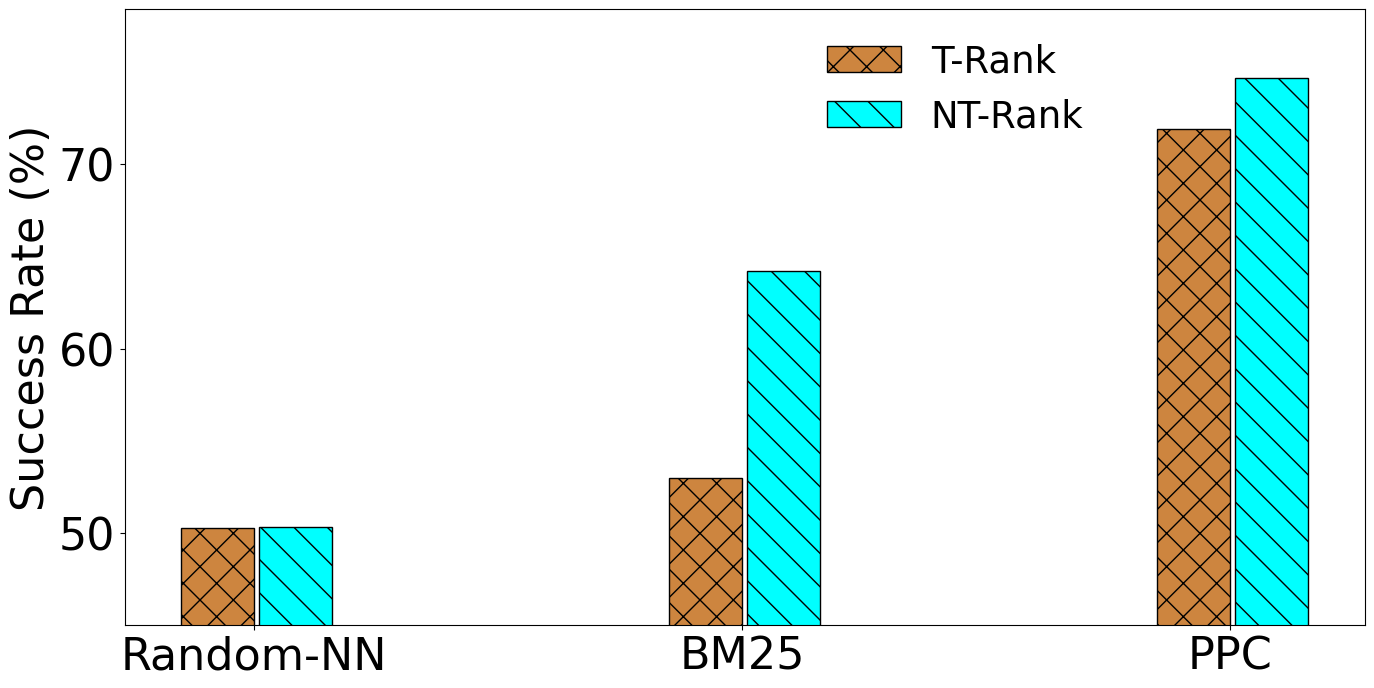}
\end{subfigure}
\caption{Completion success rate with different approaches to producing repository contexts  (RCs). \textbf{\textit{(Left)}} Impact of RC production and ordering strategies; \textbf{\textit{(Right)}} Impact of different RC retrieval methods.}
\label{fig:passage_mode_and_context_type}
\end{figure}

\textbf{Prompt Proposals Matter:}
The right side of Figure~\ref{fig:passage_mode_and_context_type} illustrates the success rate of RepoFusion using Random-NN, BM25, and PPC (refer to Section~\ref{repo_context} for details) when employing T-Rank and NT-Rank. Note that when evaluating Random-NN and BM25, we employed corresponding RepoFusion models specifically trained to accept Random-NN and BM25 contexts as inputs. The results show that using the repo context from PP~\cite{shrivastava2022repository} performs the best.

\textbf{The NT-Prior-Last Strategy is Most Effective:}
Next, we compare performances of the four different repo context production and ordering strategies that we introduced in Section~\ref{repofusion}.
The left side of Figure~\ref{fig:passage_mode_and_context_type} illustrates the success rate for the four strategies in two distinct settings: $N=32, l=768$ and $N=63, l=512$. We see that the ordered repo contexts, specifically NT-Prior-Last, NT-Rank, and T-Rank perform better than random ordering of repo contexts (T-Rand). Also, the improved performance of NT-versions when compared to the T-versions, highlights the value of presenting complete context from top prompt proposals, as opposed to displaying truncated contexts from more prompt proposals.

\begin{figure}[H]
\begin{subfigure}{.325\textwidth}
\includegraphics[width=1.0\linewidth]{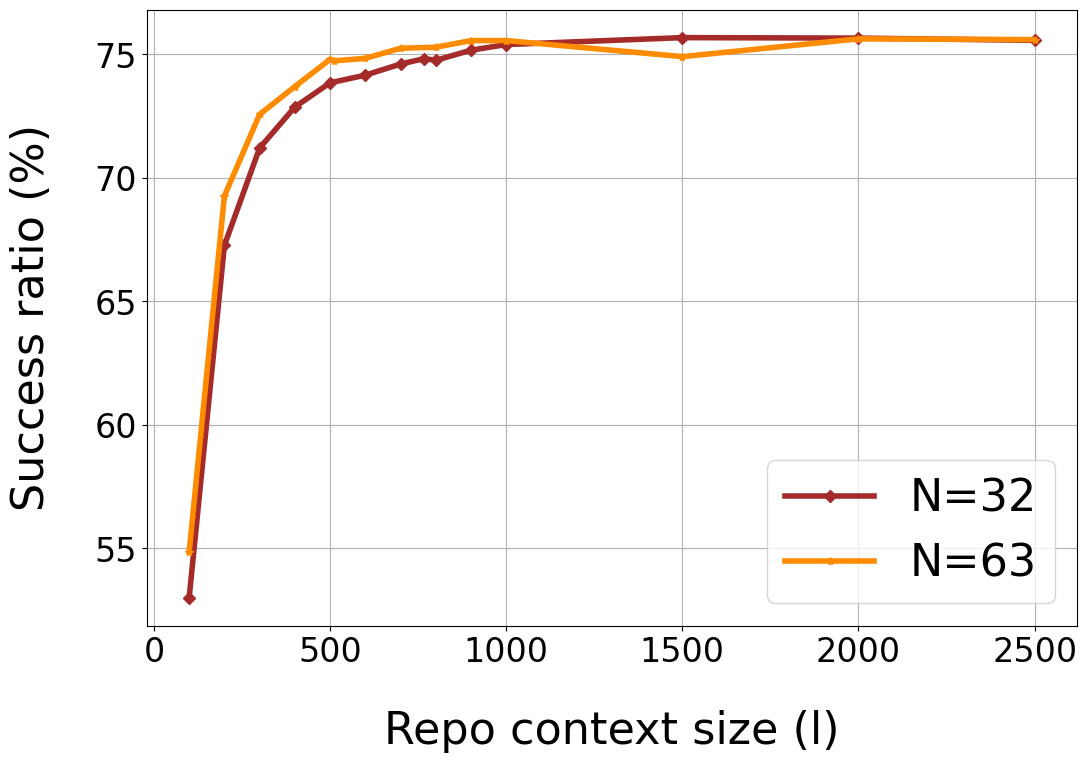}
\end{subfigure}
\hfill
\begin{subfigure}{.325\textwidth}
\includegraphics[width=1.0\linewidth]{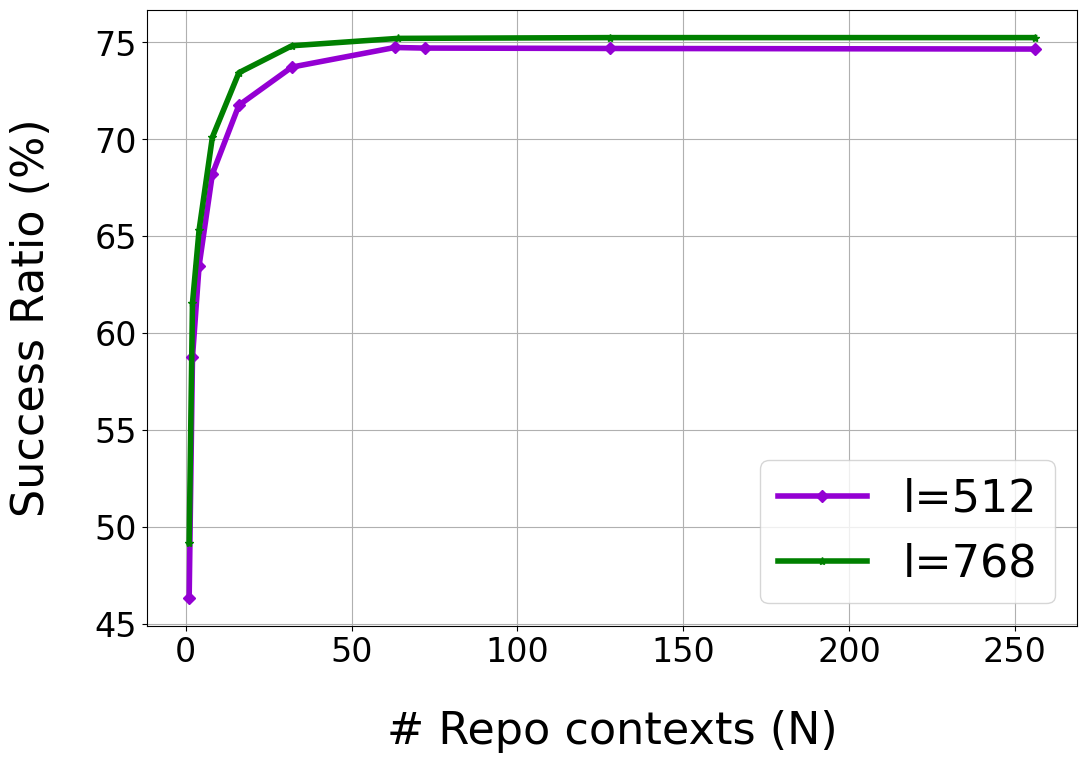}
\end{subfigure}
\hfill
\begin{subfigure}{.325\textwidth}
\includegraphics[width=1.0\linewidth]{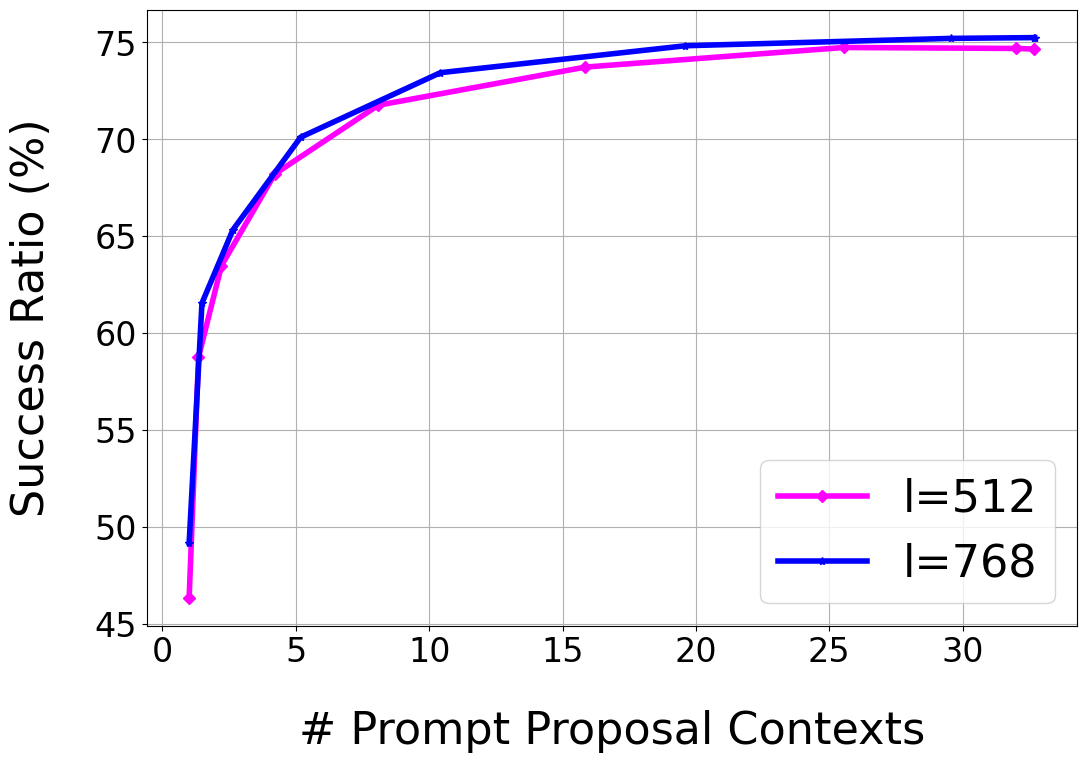}
\end{subfigure}
\caption{Completion success rate as a function of \textbf{\textit{(Left)}} the length of the individual repo context $l$; \textbf{\textit{(Middle)}} the number of repo contexts $N$; \textbf{\textit{(Right)}} the number of prompt proposal contexts that were used to produce the $N$ repo contexts.}
\label{fig:context_len_and_n_contexts}
\end{figure}

\textbf{Longer Repo Contexts are Better:} In the left side of Figure~\ref{fig:context_len_and_n_contexts}, we plot the variation of success rate with different values of the repo context length $l$. For this experiment, we used our best-performing model that was trained using NT-Prior-Last. The results indicate an improvement in the performance with the size of each repo context. However, in both cases ($N=32, N=63$), the performance reaches a saturation point as the value of $l$ increases.

\textbf{Using More Repo Contexts is Better:} 
To understand the role of multiple repo contexts, we evaluated our best-performing model with different values of $N$. We see from the middle part of Figure~\ref{fig:context_len_and_n_contexts} that the performance of RepoFusion increases upto $N=63$ when $l=512$ and up to $N=32$ with a longer length of repo context $l=768$. After this, increasing the value of $N$ doesn't lead to further improvements.

\textbf{RepoFusion Benefits from Diverse Prompt Proposals:} 
We additionally look at the success rate as a function of the average number of different prompt proposal contexts that produced the considered repo contexts for each number of repo contexts $N$. Note that the number of PPCs is less than $N$ because often one PPC yields multiple RCs. 
One can see from the right part of Figure~\ref{fig:context_len_and_n_contexts} that using many diverse PPCs was essential for getting better performance with RepoFusion.

\textbf{Finetuning the Base Model for Next Token Prediction is Important:}
Table~\ref{hole_init} shows the results of evaluating RepoFusion when the corresponding model is trained by initializing with a pretrained CodeT5-base model versus initializing with a finetuned version. We observe that while training with repo contexts enhances the performance of the base pretrained CodeT5 model (see Appendix~\ref{app:pretrained_codet5} for the performance with pretrained CodeT5), we see that in all cases, there are clear benefits of initializing the training with a model that is finetuned for code completion. We also carry out experiments to study the effect of repetition of repo contexts and not including the surrounding context. See Appendix~\ref{app:additional_results} for details.

\begin{table}[H]
\caption{Completion success rate when initialized from a pretrained vs finetuned model.}
\centering
\scalebox{0.9}{
\begin{tabular}{lcc}
\toprule
 & \textbf{Pretrained}
 & \textbf{Finetuned}\\
\midrule
\textbf{T-Rand} & 54.67$\pm$0.50 & 66.53$\pm$0.47\\  
\textbf{T-Rank} & 59.57$\pm$0.49 & 72.78$\pm$0.45\\
\textbf{NT-Rank} & 60.88$\pm$0.49 & 73.60$\pm$0.44\\
\textbf{NT-Prior-Last} & 61.91$\pm$0.49 & 74.82$\pm$0.43\\
\bottomrule
\end{tabular}}
\label{hole_init}
\end{table}

\section{Related Work}\label{related_work}

\textbf{Information from Outside the Current File:} In the context of source code, harnessing information beyond the current file has been found to be useful. \citet{n-gram} utilizes a nested n-gram model with a locality-based cache encompassing all directories in the repository. To capture the structure of the repository, \citet{pashakhanloo2022codetrek, pashakhanloo2022learning} convert it into a relational database and propose a graph-walk mechanism whereas, \citet{API} incorporates the API-dependency graph in its LSTM-based code model. While training the kNN-LM~\citep{Khandelwal2020Generalization}, \citet{xu2022capturing} incorporates three types of binary variables corresponding to the presence or absence of similar local and global hierarchy. \citet{class_name} leverages the parent class to generate comments for the child class. 

\textbf{Repository-level Context for Inference in LLMs:} 
\citet{shrivastava2022repository} proposes RLPG, a classifier that selects a prompt proposal based on the target hole and utilizes the context from the chosen prompt proposal and prior context to prompt Codex~\cite{codex}. Similarly, RepoCoder~\cite{repocoder} iteratively refines the prediction of the target hole by injecting the previous predictions from the LLM in addition to the retrieved context and prior context in the input prompt. In this work, we build upon the insights gained by \citet{shrivastava2022repository} regarding the utilization of repository-level information during inference. We extend their findings to various configurations involving different code language models (LLMs), considering a range of context lengths and sizes. Additionally, our framework is trained with context from the repository and learns to effectively leverage multiple relevant contexts sourced from the repository.

\textbf{Retrieval-augmented Code Models:} In recent studies~\cite{zhou2023docprompting, parvez-etal-2021-retrieval-augmented, lu2022reacc,zan-etal-2022-language, cocomic, borgeaud2022improving}, attempts have been made to enhance code LLMs by augmenting them with a sparse or dense retrieval mechanism that returns API documentation or relevant code snippets from the repository. The prompt proposals~\cite{shrivastava2022repository} used in our work along with BM25 and Random-NN share similarities with these retrieval mechanisms. Note that RepoFusion is independent from the specific retrieval mechanisms employed and thus can seamlessly learn to integrate multiple retrieved contexts, even from different retrieval mechanisms.

\section{Discussion}\label{conclusion}
 
\textbf{Limitations} 
RepoFusion has a limitation in terms of computation scalability as it exhibits linear scaling with respect to the number of repo contexts $N$, leading to slower inference times for larger values of $N$. One possible solution to address this issue is to leverage FiDO~\cite{deJong2022FiDOFO}, an optimization technique for FiD that enables faster inference. Deploying RepoFusion, similar to any other code LLMs, requires careful consideration~\cite{codex}. The generated code can often be challenging to understand or debug, resulting in developers spending significant time editing and revising the code~\cite{vaithilingam2022expectation, mozannar2022reading, barke2023grounded, bird2022taking}. There can be instances where the generated code is less secure, posing potential risks~\cite{pearce2021empirical}. Moreover, excessive dependence on these models can result in situations where users overlook errors in their code~\cite{al2022readable} or become overly self-assured, leading to the introduction of mistakes~\cite{perry2022users}.

\textbf{Conclusions and Future Work}
We propose RepoFusion, a framework that allows training code models with multiple relevant contexts from the repository. By employing RepoFusion in experiments focused on single-line code autocompletion, we highlight the notable enhancements in performance attained through training smaller models with repository context, surpassing the results of training larger models without such context. RepoFusion, in combination with the Stack-Repo dataset, opens up exciting avenues for future research in the field of smaller retrieval-augmented LLMs for code. We believe our method can also extend to other code-related tasks such as bug repair, the merging of pull requests, and software documentation/tutorial writing.

\section*{Acknowledgements}
The authors would like to thank ServiceNow Research for providing compute resources required for this project. We would like to acknowledge the authors of \citet{fid, shrivastava2022repository} for making their code publicly available, which served as a foundation for building our work. We would also extend our gratitude to the authors of The Stack~\cite{kocetkov2022stack} for releasing their data and \citet{codegen, code-t5, starcoder, santacoder} for releasing their models. We would like to thank Hugo Larochelle and Daniel Tarlow who contributed to initial discussions about combining multiple repository contexts that served as the motivation for this work. We would like to thank Sébastien Paquet for feedback on the draft that helped us improve our writing. Lastly, we would like to thank Sami Turcotte for helping with the figures used in the paper.

\bibliographystyle{plainnat}{
\bibliography{neurips_2023}
}


\newpage

\appendix

\section{Details on Stack-Repo}\label{app:stack-repo}
We have made our dataset available at the link:~\url{https://huggingface.co/datasets/RepoFusion/Stack-Repo}. The details of the license can be found in the Licensing Information section of the page. Stack-Repo consists of 200 near-deduplicated Java repositories (see Table~\ref{tab:dataset_stats} of the main paper for details). For each repository within a split (train, validation and test), we provide all files arranged in the directory structure within the repository along with three \code{.json} files that contain the PP, BM25 and RandomNN repo-contexts. One row of the \code{.json} file corresponds to a target hole consisting of the location of the target hole, the target hole as a string, the surrounding context as a string and a list of repo-contexts as strings.

\section{Implementation Details}\label{app:implementation}
\subsection{Finetuning CodeT5}\label{app:finetune}
As described in Section 3.2 of the main paper, to serve as a better initialization of RepoFusion (also served as a baseline) we finetuned a CodeT5-base model (220M parameters) with an input context length of 512 tokens using the CodeT5 tokenizer. We used an Adam optimizer with Decoupled Weight Decay Regularization~\cite{DBLP:conf/iclr/LoshchilovH19} with weight decay of 0.05 and a learning rate of 4e-05. In addition, we used a linear scheduler with 100 warm-up steps, a dropout of 0.1, and gradient clipping with a max gradient norm of 1.0. To serve as a baseline, we also finetuned a CodeT5-large model (770 M parameters) with an input context length of 512. We used the same set of hyperparameters for this as mentioned before except that we used a learning rate of 1e-4. The training was carried out on 2 NVIDIA A100 GPUs with a memory of 80GB each and a batch size of 32 per GPU for the CodeT5-base model. For CodeT5-large we used 4 A100 GPUs with memory of 80GB each and a batch size of 12 per GPU. The evaluation run was carried out on a single 32GB V100 GPU with a batch size of 32 for CodeT5-base and 48 for CodeT5-large.

\subsection{Training RepoFusion}
We use the 220M parameter CodeT5-base~\cite{code-t5} encoder-decoder model as our base code model for RepoFusion. Our RepoFusion implementation was heavily built on top of the code released by \citet{shrivastava2022repository}\footnote{https://github.com/shrivastavadisha/repo\_level\_prompt\_generation (MIT License)}, as well as the code released by \citet{fid}\footnote{https://github.com/facebookresearch/FiD (Creative Commons Attribution-NonCommercial 4.0 International Public
License)}. The former was used to obtain repo contexts and the latter was used for the FiD architecture.  

Our best RepoFusion model was obtained by initializing the training from a finetuned CodeT5-base checkpoint (see Section~\ref{app:finetune} for details). The repo contexts used the NT-Prior-Last strategy (see Section 2.3 of the main paper for details) with 32 PP repo contexts each of size 768 tokens ($N=32, l=768$). Similar to~\citet{fid}, we format each repo context with special tokens to mark the beginning of the surrounding context and the repo context, as well as for the name of the repo context (which is the same as the name of the PP taken from~\cite{shrivastava2022repository}). We used \code{hole\_context:} as prefix for the surrounding context, \code{rule\_context:} as a prefix for PP repo context, and \code{rule\_name:} as prefix for PP repo context name. We used Adam~\cite{adam} optimizer with a learning rate 1e-5 and a warmup linear scheduler with 5000 warmup steps. We used gradient clipping with norm 1.0 and batch size of 1. Training was carried out on 2 NVIDIA A100 GPUs with memory of 80GB each. Each evaluation run was carried out on a single 32GB V100 GPU. 

The BM25 and Random NN versions of RepoFusion were obtained by using the same training hyperparameters as above and initialized from the same finetuned CodeT5 checkpoint except that we found that a learning rate of 2.5e-5 and the setting $N=63, l=512$ works the best. As before, we used NT-Prior-Last strategy and a prefix only for the surrounding context and no prefixes for repo contexts. The RepoFusion model that was initialized from a pretrained CodeT5-base version was obtained by using the same set of training hyperparameters as our best RepoFusion model but a learning rate of 1e-4 worked the best.

We release the RepoFusion models as well as the finetuned CodeT5 models at~\url{https://huggingface.co/RepoFusion/trained_checkpoints}.

\subsection{Retrieval Mechanisms}\label{app:implementation_retrieval_mechanisms}
The BM25 repo contexts were obtained using the Okapi BM25 implementation with default parameters given by the pip package~\code{rank-bm25 0.2.2}\footnote{https://pypi.org/project/rank-bm25/}. The BM25 scores are calculated with the surrounding context being the query and full context from other files in the repository being the search documents. 
Random NN repo contexts used the procedure followed by \citet{shrivastava2022repository} using CodeBERT~\cite{feng2020codebert} to obtain the representations (See Appendix C.3 for details).

\subsection{Other Baselines}
We used the models available on Hugging Face hub, i.e.\ \href{https://huggingface.co/Salesforce/codegen-2B-multi}{Codegen-2B-multi}, \href{https://huggingface.co/Salesforce/codegen-6B-multi}{CodeGen-6B-multi}, \href{https://huggingface.co/Salesforce/codegen-16B-multi}{CodeGen-16B-multi}, \href{https://huggingface.co/bigcode/santacoder}{SantaCoder} and \href{https://huggingface.co/bigcode/starcoder}{StarCoder}. We used special FIM tokens, i.e., \code{<fim-prefix>} for pre context,  \code{<fim-suffix>} for post context and \code{<fim-middle>} to prompt for completing the target hole. Each of these models used the recommended tokenizers and completion length of 128 tokens.

\section{Additional Results}\label{app:additional_results}

\subsection{Effect of Repetition}
In order to further assess the significance of diverse repo contexts, we conducted an analysis by repeating a PPC multiple times and using each repetition as a separate repo context. One can see from the right side of Table~\ref{fig:repetition} that repeating the context from a single prompt proposal (prior, post, randomly chosen PP) has a negative impact on performance compared to using different repo contexts from multiple prompt proposals.

\begin{table}[H]
\caption{Completion success rate with repetiting different types of PPCs multiple times.}
\centering
\scalebox{1.0}{
\begin{tabular}{lc}
\toprule
 & \textbf{\begin{tabular}[c]{@{}c@{}} Success Rate(\%) \end{tabular}}\\
\midrule
\textbf{Rand} & 37.18$\pm$0.48\\
\textbf{Prior} & 50.69$\pm$0.50\\   
\textbf{Post} & 54.64$\pm$0.50\\
\textbf{NT-Rank} & 71.92$\pm$0.45\\
\bottomrule
\label{fig:repetition}
\end{tabular}}
\end{table}

\subsection{Appending Surrounding Context}
Table~\ref{fig:appending_surr_context} shows the performance of RepoFusion when we do not append the surrounding context to each repo context. We see that the performance drops significantly for all strategies when compared to when the surrounding context is appended. It should be noted that for these experiments, we used our best RepoFusion model that is trained to take the concatenation of surrounding context and repo context as input. It is highly likely that a RepoFusion model trained to not append the surrounding context would suffer from much less performance drop.
\begin{table}[H]
\caption{Completion success rate with and without appending surrounding context.}
\centering
\scalebox{1.0}{
\begin{tabular}{lcc}
\toprule
 & \textbf{\begin{tabular}[c]{@{}c@{}} without\\ Surrounding Context \end{tabular}}
 & \textbf{\begin{tabular}[c]{@{}c@{}} with\\ Surrounding Context \end{tabular}}\\
\midrule
\textbf{T-Rand} & 13.89$\pm$0.35 & 66.53$\pm$0.47\\  
\textbf{T-Rank} & 25.06$\pm$0.43 & 72.78$\pm$0.45\\
\textbf{NT-Rank} & 15.57$\pm$0.36 & 73.60$\pm$0.44\\
\textbf{NT-Prior-Last} & 17.18$\pm$0.38 & 74.82$\pm$0.43\\
\bottomrule
\label{fig:appending_surr_context}
\end{tabular}}
\end{table}

\subsection{Performance of Pretrained CodeT5}\label{app:pretrained_codet5}
Table~\ref{tab:pretrained_codet5} shows the performance on the test set when we directly use the pretrained CodeT5-base and CodeT5-large models. For these experiments, we use the special token \code{<extra\_id\_0>} to prompt the completion of the target hole. We see that the performance of these pretrained models is quite low, thereby creating the need to finetune these models on Java repositories on the next-token prediction objective. We see from the top section of Table 2 in the main paper that the finetuning helps a lot. 

\begin{table}[H]
\centering
\caption{Completion success rate on the test set for pretrained CodeT5.}
\scalebox{1.0}{
\begin{tabular}{@{}cccccc@{}}
\toprule
\textbf{\begin{tabular}[c]{@{}c@{}}Model\end{tabular}} & 
\textbf{\begin{tabular}[c]{@{}c@{}} Size \\(\#params)\end{tabular}} &
\textbf{\begin{tabular}[c]{@{}c@{}} Effective \\ context length \end{tabular}} &
\textbf{\begin{tabular}[c]{@{}c@{}} Context \\ type \end{tabular}} &
\textbf{\begin{tabular}[c]{@{}c@{}} Success Rate \\(\%)\end{tabular}} \\
\midrule
CodeT5-base & 0.22B & 512 & prior & 2.42 (0.04)\\
CodeT5-base & 0.22B & 2048 & prior & 3.94 (0.05)\\
CodeT5-large & 0.77B & 512 & prior &  4.56 (0.05)\\ 
CodeT5-large & 0.77B & 2048 & prior &  9.51 (0.07)\\
\bottomrule
\end{tabular}
\label{tab:pretrained_codet5}}
\end{table}

\end{document}